%% file: main.tex
\pdfoutput=1
\documentclass[11pt]{article}

\usepackage[final]{acl}

\usepackage{times}
\usepackage{latexsym}

\usepackage[T1]{fontenc}
\usepackage[utf8]{inputenc}

\usepackage{microtype}

\usepackage{inconsolata}

\usepackage{graphicx}

\usepackage{multirow}
\usepackage{multicol}
\usepackage[normalem]{ulem}
\usepackage{paralist}
\usepackage{booktabs}
\usepackage{comment}
\usepackage{amsmath}
\usepackage{soul}

\useunder{\uline}{\ul}{}

\title{Selective Self-Rehearsal: A Fine-Tuning Approach to Improve Generalization in Large Language Models}

\author{
 \textbf{Sonam Gupta\textsuperscript{1}},
 \textbf{Yatin Nandwani\textsuperscript{1}},
 \textbf{Asaf Yehudai\textsuperscript{1}},
 \textbf{Mayank Mishra\textsuperscript{1}},
\\
 \textbf{Gaurav Pandey\textsuperscript{1}},
 \textbf{Dinesh Raghu\textsuperscript{1}},
 \textbf{Sachindra Joshi\textsuperscript{1}}
\\
\\
 \textsuperscript{1}IBM Research
\\
 \small{
   \textbf{Correspondence:} \{sonam.gupta7, yatin.nandwani\}@ibm.com
 }
}

\begin{document}
\maketitle
\begin{abstract}
\input{abstract}
\end{abstract}
\section{Introduction}
\input{intro}
\section{Proposed Method}
\input{method}

\section{Experimental Setup}

\input{expt}

\section{Results and Discussion}
\label{sec:results}
\input{results}

\section{Related Work}
\input{rel}

\section{Conclusion}
\input{conclusion}
\section{Limitations}
\input{limitation}

%\section*{Acknowledgments}

\bibliography{custom}

\newpage
\appendix

\section{Appendix}
\label{sec:appendix}
\input{appendix}

\end{document}

%% file: abstract.tex
Fine-tuning Large Language Models (LLMs) on specific datasets is a common practice to improve performance on target tasks. However, this performance gain often leads to overfitting, where the model becomes too specialized in either the task or the characteristics of the training data, resulting in a loss of generalization. This paper introduces Selective Self-Rehearsal (SSR), a fine-tuning approach that achieves performance comparable to the standard supervised fine-tuning (SFT) while improving generalization.
SSR leverages the fact that there can be multiple valid responses to a query. By utilizing the model's correct responses, SSR reduces model specialization during the fine-tuning stage. SSR first identifies the correct model responses from the training set by deploying an appropriate LLM as a judge. Then, it fine-tunes the model using the correct model responses and the gold response for the remaining samples.
The effectiveness of SSR is demonstrated through experiments on the task of identifying unanswerable queries across various datasets. The results show that standard SFT can lead to an average performance drop of up to $16.7\%$ on multiple benchmarks, such as MMLU and TruthfulQA. In contrast, SSR results in close to $2\%$ drop on average, indicating better generalization capabilities compared to standard SFT.

%% file: intro.tex
\newcommand{\question}{{\small {Do I need insurance to register my car?}}}
\newcommand{\goldanswer}{{\small {Yes. New York law requires that you have auto liability insurance coverage.}}}
\newcommand{\prediction}{{\small {Yes, according to the context provided, you need New York State issued automobile liability insurance coverage to register a vehicle in New York State.}}}

\newcommand{\mistral}{Mistral-7B-Instruct-v0.2}

\input{intro_example}

\begin{figure}
    \centering
    \includegraphics[width=\linewidth]{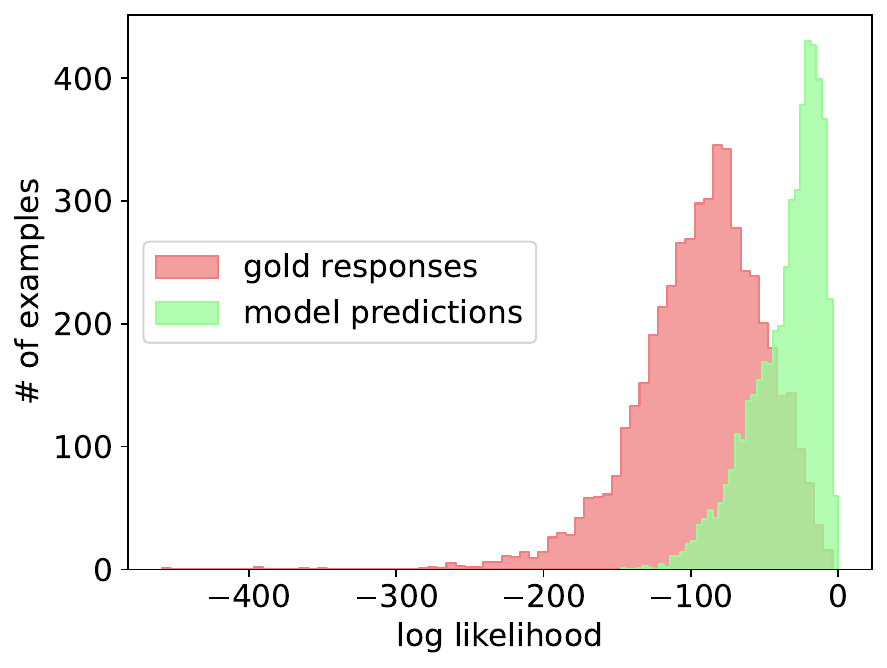}
    \caption{Histogram of the log probability assigned by \mistral\ to the gold responses and its own predictions. The distribution is based on 5,000 examples from the MD2D training data.}
    \label{fig:prob_hist}
\end{figure}

% Problem Statemnt: genral model somtime needs adjustments

Large Language Models (LLMs) have made remarkable progress in recent years, demonstrating impressive capabilities across a wide range of tasks, including question-answering \cite{rajpurkar2016squad}, summarization \cite{nallapati2016abstractive}, and more \cite{brown2020language}. This advancement has led to the adoption of LLMs in various real-life applications, such as customer support \cite{xu2017new} and code assistance \cite{chen2021evaluating}. However, adapting these models to specialized domains and tasks often requires adjustments to meet the specific unique needs of model designers. For example, a designer of a customer support agent may want the model to abstain from answering questions that are unanswerable, off-topic, or potentially unsafe.

Current approaches to address this challenge include prompt engineering and fine-tuning with task-specific data. Prompt engineering involves guiding the model's behavior through instructions and few-shot in-context examples without altering its weights, allowing it to retain its original capabilities. However, this method may lead to sub-optimal performance on the target task \cite{stiennon2020learning}. Fine-tuning, on the other hand, can better align the model with the desired behavior \cite{peters2019tune}, but may reduce the model's generality. Our work aims to follow the fine-tuning approach while aiming to maintain the model's general capabilities.

Supervised Fine-Tuning (SFT) typically relies on gold responses for training. However, for instruction-tuned models, we observe two key issues: 1) many model responses, while differing from gold responses, are still satisfactory, and 2) the distribution of gold responses often diverges significantly from the model's own response distribution.
For example, consider the example in Table \ref{tab:intro-example}. The base model, \mistral, assigns a log probability of $-109.9$ to the gold answer. When prompted with the same question, the model generated prediction has the same information as gold, but its log probability is $-2.4$. This phenomenon is common in many generation tasks, where different responses can convey the same meaning with very different values of log likelihood. Moreover, as illustrated in Figure \ref{fig:prob_hist}, there is a clear gap between the distributions of gold responses and the model's learned responses on a set of 5000 examples. This indicates that model-generated responses can be valid and closer to the model's own distribution, while gold responses may be further apart. Consequently, training exclusively on gold responses can lead to a drift from the original distribution, compromising the model's generality.

To address these issues, we propose Selective Self-Rehearsal (SSR), a fine-tuning approach that utilizes model-generated answers for a subset of the training dataset to adapt the model to desirable behaviours while maintaining generalization. 
SSR fine-tunes the model on its own generated output for cases where it behaves desirably and on gold output for the remaining data. This approach allows the model to learn from its own successes while still benefiting from human-labeled data when needed.

To showcase our method, we focus on content-grounded QA/conversation, where the model needs to respond to user queries based on provided content or identify the query as 'unanswerable' and respond appropriately. In this context, the general ability is answering 'answerable' questions, and the required modification is correctly identifying 'unanswerable' questions. Our objective is to teach the model to identify 'unanswerable' queries while retaining its original capabilities, including responding to 'answerable' queries.

Our extensive experiments on multiple unanswerability datasets from different domains and styles demonstrate the effectiveness of our simple yet powerful method. To show that SSR generalizes better and retains the base model's capabilities, we evaluate the fine-tuned model on multiple datasets for the same and different tasks and domains. For our evaluation on the benchmarks MMLU \cite{mmlu}, TruthfulQA \cite{truthfulqa}, and Hellaswag \cite{hellaswag}, we observe that standard SFT results in up to a $16.7\%$ average drop in performance over these benchmarks, while SSR results in close to $2\%$ drop on average, demonstrating better generalization capabilities of SSR over standard SFT.

%% file: intro_example.tex
\begin{table}[t]
\small
\begin{tabular}{p{0.2\columnwidth}@{}|p{0.75\columnwidth}@{}}
\toprule
\textit{Document} & To register a vehicle in New York State you must have New York State issued automobile liability insurance coverage ... \\ \cmidrule(l){1-2}
\textit{Question} & \question \\ \cmidrule(l){1-2}
\textit{Gold} & \goldanswer  \par \sethlcolor{pink} \hl{log probability = -109.9} \\ \cmidrule(l){1-2}
\textit{Model} \par \textit{Prediction} & \prediction  \par \sethlcolor{lime} \hl{log probability = -2.4} \\ \bottomrule
\end{tabular}
%}
\caption{An example from the MultiDoc2Dial, along with \mistral's prediction.}
\label{tab:intro-example}
\end{table}

%% file: method.tex
\newcommand{\model}{\mathcal{M}}
\newcommand{\basemodel}{\mathcal{M}_{\theta_0}}

\newcommand{\task}{\mathcal{T}}
\newcommand{\traindata}{\mathcal{D}}
\newcommand{\inp}{x}
\newcommand{\outp}{y}
\newcommand{\loss}{\mathcal{L}}
\newcommand{\sftloss}{\mathcal{L}_{SFT}}
\newcommand{\prob}{Pr}
\newcommand{\pred}{\hat{y}}
\newcommand{\etc}{\emph{etc.}}
\newcommand{\goodsubset}{\mathcal{R}}
\newcommand{\goldsubset}{\mathcal{G}}

\newcommand{\ssrloss}{\mathcal{L}_{SSR}}

\begin{figure*}
    \centering
    \includegraphics[width=\linewidth]{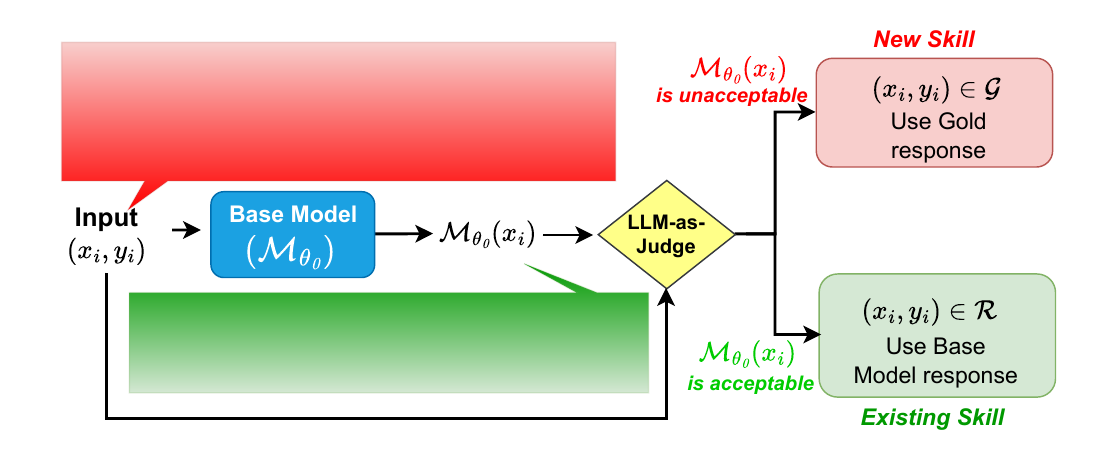}
    \caption{An overview of our proposed approach. In the example, the document and question are part of the input $\inp_i$, and the response is the output $\outp_i$. The llm-judge decides whether the base model output $\basemodel(\inp_i)$ is acceptable or not. If yes, then we use it for loss computation (subset $\goodsubset$); otherwise, we use $\outp_i$ (subset $\goldsubset$). See eqn. \ref{eqn:ssrloss}.}
    \label{fig:overview}
\end{figure*}

Let $\model_\theta$ parameterized by $\theta$ be a given large language model. 
Let $\theta = \theta_0$ be the given model weights obtained after pre-training and instruction tuning the model.
We refer to $\model_{\theta_0}$ as the base model.
Let us assume that we cannot access the pretraining and instruction fine-tuning datasets.
Further, let $\task$ be the new task that we wish to teach the model $\model_\theta$, and let $\traindata = \{(\inp_i, \outp_i) | i = 1 \ldots N\}$ be the corresponding dataset that we may use to teach the new task to the model.
In standard Supervised Fine-Tuning (SFT), we backpropagate through the standard Cross Entropy loss over the training dataset, computed as: 
\begin{equation}
    \sftloss(\traindata) = 
    - \sum\limits_{i=1}^{N} \log \prob_\theta(\outp_i | \inp_i)
\end{equation}
Here, $\prob_\theta$ is the conditional probability assigned by the model $\model_\theta$.

Now, let us assume that for an input $\inp_i$, $\pred_i = \model_{\theta_0}(\inp_i)$ is the base model's prediction.
We know that in many applications of NLP, e.g. machine translation,  content-grounded conversations, summarization, reading comprehension \etc, an input $\inp$ can have multiple
correct answers, and it may suffice to generate any one of them.
Given that the model has already been instruction-tuned on a variety of tasks, it is quite likely that the prediction $\pred_i=\model_{\theta_0}(\inp_i)$ of the base model is as good as the given gold answer $\outp_i$.
If that is the case, we ask the research question: which of the two outputs, $\pred_i$ or $\outp_i$, should be used to compute the loss?
\cite{nandwani2020neural} define such a setup where there are multiple correct solutions for a given input as 1oML (one of many learning) and propose various strategies to handle it, albeit for combinatorial problems.
Taking inspiration from it, we hypothesize that using $\pred_i$ instead of $\outp_i$ to compute the loss in such a scenario regularizes the model and helps in tackling catastrophic forgetting of the skills acquired during the instruction tuning phase.
We note that the standard practice of regularization via replay buffer \cite{hayes2020remind},
which involves mixing a subset of instruction-tuning dataset with the given task-specific data $\traindata$ is not always feasible as the instruction-tuning dataset may not be available.

Formally, let $\goodsubset \subseteq \traindata$ be a subset of the given training dataset such that for $(\inp_i,\outp_i) \in \goodsubset$, base model's prediction $\pred_i = \basemodel(\inp_i)$ is as good as the given gold output $\outp_i$. Let $\goldsubset = \traindata - \goodsubset$ be the remaining dataset.
In our proposed Selective Self-Rehearsal technique, we compute the loss as follows:
\begin{align}
    \label{eqn:ssrloss}
    \ssrloss(\traindata) = 
    -  & \sum\limits_{(\inp_i,\outp_i) \in \goodsubset} \log \prob_\theta(\pred_i | \inp_i) \ \ -  \notag \\
     & \sum\limits_{(\inp_i,\outp_i) \in \goldsubset} \log \prob_\theta(\outp_i | \inp_i)
\end{align}

Now, the question arises: how do we know if the base model's prediction $\pred_i=\basemodel(\inp_i)$ for an input is as good as the corresponding gold response $\outp_i$?
To answer this, we may either use a heuristic to measure the goodness of the prediction or, alternatively, as prevalent these days, we may prompt a powerful LLM, such as Mixtral-8x7B \cite{mixtral} or GPT-4 \cite{gpt4}, to compare $\pred_i$ with $\outp_i$ and evaluate if $\pred_i$ is as good as $\outp_i$ or not.
See fig. \ref{fig:overview} for an overview of our approach.

%% file: expt.tex
\noindent \textbf{Task:} Our experiments aim to compare the proposed SSR method with the standard SFT for teaching a new task to the LLM. We focus on teaching the task of content-grounded QA/conversation. In this task, the LLM must respond based on the information present in the provided document. If the document doesn't contain the information necessary to respond, then the LLM must refrain from responding and inform the user that it can't find the information in the provided document.  We observe that the base LLM generates acceptable responses when the document contains the answer. However, it hallucinates when the question can't be answered from the provided document. This perfectly fits the premise for SSR method -- the base LLM is good at answering the `\textit{answerable}' questions, while it  needs to learn to refrain from answering for the `\textit{unanswerable}' queries. 

\vspace{0.2ex}
\noindent \textbf{Datasets:} 
To fine-tune a base LLM, we use two publicly available content-grounded QA/conversation datasets: (1) natural questions (NQ) \cite{nqdataset}, and (2) MultiDoc2Dial (MD2D) \cite{multidoc2dial}. NQ is a content-grounded QA dataset. \citet{slobodkin2023curious} augment the NQ dataset with unanswerable queries, making it suitable for our setup. Here, the grounding content consists of a single paragraph, and the gold answers are short phrases. MD2D is a multi-turn document-grounded conversational dataset. 
%This dataset doesn't contain unanswerable turns, so we augment it with unanswerable turns.
This dataset lacks unanswerable turns, so we augment it by adding them.
As each conversation in the dataset is grounded on multiple documents, we identify the turn where the document changes and replace the document with an incorrect one to synthesize unanswerable turns systematically.

To study the ability of the fine-tuned model to generalize to other datasets for the same task, we test the model fine-tuned using each of the above two datasets on MuSiQue \cite{trivedi2022musique} dataset. We use the augmented version \cite{slobodkin2023curious} of the dataset, which has unanswerable questions. 
MuSiQue is a content-grounded multi-hop reasoning QA dataset. This dataset helps us evaluate the LLM's ability to generalize to domains unseen during train. See Table \ref{table:test-data-statistics} for the statistics of all three test datasets.

To test the finetuned model's ability to retain the base model's capabilities,  we evaluate the fine-tuned models on several standard benchmarks such as MMLU \cite{mmlu}, Truthful-QA \cite{truthfulqa}, GSM8k \cite{gsm8k} and Hellaswag \cite{hellaswag}.

\input{data_stats}

\vspace{0.2ex}
\noindent \textbf{Evaluation Metrics:} Following \citet{adlakha-etal-2024-evaluating}, we use the token level recall between the predicted response and the gold response to measure the quality of the responses generated for content-grounded QA/conversation. 
As our datasets have both answerable and unanswerable classes, we penalize an example for predicting an answerable query as unanswerable, and vice-versa, by assigning it a recall of $0$.
For an unanswerable query, if the model correctly predicts it as unanswerable, we assign it a perfect recall of $1$.
We use classification accuracy to measure the model's ability to classify between the two classes (answerable vs unanswerable). 
We initially relied on string-matching heuristics to design a rule-based classifier. For example, it would search for strings such as "I don't know", "unanswerable", etc, in a response and classify it as `unanswerable' if such a string is present in it. 
However, we found many examples where it failed as the base model may say `I don't know' in many ways and it may not be possible to cover it all using rules. 
Hence, we decided to create a prompt and employ Mixtral-8x7B \cite{mixtral} as a judge, prompting it to classify a response as either answerable or unanswerable.
To measure the efficacy of the prompt, two authors manually annotated 175 responses and computed the accuracy of the two systems. 
The heuristics achieved an accuracy of $86.6\%$ whereas our llm-judge attained an accuracy of $96\%$, and hence we decided to proceed with the llm-judge.
See Appendix \ref{app:llm-as-judge-prompt} for the exact prompt.

\vspace{0.2ex}
\noindent \textbf{Human Evaluation:} We also perform a human evaluation to the study the performance of SSR over other approaches. We measure \textit{relevance}, the ability to generate relevant responses for the given dialog context and the provided document on a Likert scale (0-4) \cite{likert1932technique}. The human judges were asked to assign a score of $0$ when a model refrains from answering an answerable query or when a model answers an unanswerable query (see appendix \ref{app: human_study}). We picked a model fine-tuned using MD2D, and randomly sampled 50 questions from the MD2D testset to measure in-domain performance, and 50 samples from the MusiQue testset to measure out-domain performance. For each sample, we collect annotations from two in-house human judges who are in our organization's payroll. Both judges are undergraduates with a background in NLP/ML.

\vspace{0.2ex}
\noindent \textbf{Base model and Baselines:} 
We experiment with Mistral-instruct-v2 (7B)\cite{mistral} as our base model. It performs well whe prompted to answer based on provided document, given its an answerable query. But it is not great at refraining from answering when the query is unanswerable. We use two baselines: (1) prompting the base model (see Appendix \ref{app:task-prompts} for the exact prompt), and (2) Supervised Fine-Tuning (SFT).

For both SFT and SSR, we use Low-Rank Adaptation (LoRA) \cite{hu2022lora} with a rank of 4, a scaling factor of 8 and a dropout of 0.1. Please see appendix \ref{app:train_detail} for more details.

%% file: data_stats.tex
\begin{table}
\centering
\begin{tabular}{lcc}
\hline
        & \multicolumn{1}{l}{Answerable} & \multicolumn{1}{l}{Unanswerable} \\ \hline
MD2D    & 5653                                    & 3609                                      \\
NQ      & 3489                                    & 7719                                      \\
MuSiQue & 1950                                    & 1316                                      \\ \hline
\end{tabular}
\caption{Number of answerable and unanswerable instances present in test set of various datasets.}
\label{table:test-data-statistics}
\end{table}

%% file: results.tex
Our experiments evaluate three research questions. 

\begin{compactenum}
\item \emph{In-Domain Performance}: How does SSR perform compared to baselines when fine-tuned and evaluated on the same dataset?
\item \emph{Out-Domain Performance}: How does SSR perform compared to baselines when evaluated on the datasets unseen during train?
\item \emph{Generalization}: How well does SSR retain the inherent capabilities of the base model post fine-tuning?
\end{compactenum}
\subsection{In-Domain Performance}
\label{subsec:same_task_same_domain}

\input{in_domain_recall_acc}

Table \ref{tab:in_domain_recall_acc} reports our modified recall and classification accuracy of Mistral-Instruct-v2-7b finetuned over MD2D and NQ datasets. Here, we evaluate the base, SFT, and SSR models over the test set corresponding to the training dataset.
To assess the base model's capability to generate responses for answerable queries, we also report token--recall (T.Recall(AA)) only for those answerable queries where the model generates a response instead of refraining from answering (\textbf{A}nswerable queries classified as \textbf{A}nswerable).
We first observe that the base model is good at answering the answerable questions (achieves good T.Recall(AA)) but struggles to identify when not to respond (poor classification accuracy). 
Hence, this is the skill we would like the model to learn by fine-tuning without forgetting its ability to generate good answers.
We observe that both SFT and SSR techniques for fine-tuning result in a model that is able to identify unanswerable queries equally well (similar accuracy). 
However, we observe that SSR retains the original model's ability to answer the questions, whereas token recall for the SFT model drops drastically compared to the base and SSR models.
As a result, SSR achieves the best overall performance as quantified by our modified recall metric.

\vspace{0.2ex}
\noindent \textbf{Human Evaluation:}
Table \ref{tab:human-eval} reports the human evaluation results on test samples from MD2D using the model fine-tuned on MD2D. We see that both SFT and SSR have been able to surpass the score of the simple prompting approach. We also see that the SFT approach is a bit better than SSR on the in-domain setup. 
We get moderate inter-annotator agreement ($\tau=0.34$) using Kendall's Tau.
The agreement is moderate as MD2D is conversational, and there are many possible ways to respond to the user. Some annotators prefer one style of response over others, \textit{e.g.}, some like short answers and others prefer a more detailed answer.

\begin{table}[]
\resizebox{\columnwidth}{!}{
\begin{tabular}{l|ccc}
\toprule
        & \textbf{Prompt} & \textbf{SFT} (MD2D)    & \textbf{SSR} (MD2D )   \\ \cmidrule{2-4}
\textbf{MD2D}    & 2.60   & \textbf{3.03} & 2.95          \\
\textbf{MuSiQue} & 2.47   & 1.83          & \textbf{2.78} \\ \bottomrule
\end{tabular}}
\caption{Human evaluation of models fine-tuned using MD2D on in domain (MD2D) and out-domain (MusiQue) datasets.}
\label{tab:human-eval}
\end{table}

\subsection{Out-domain Performance}
This experiment aims to demonstrate that SSR achieves better generalization than SFT. To do so, we train the model on one dataset and evaluate its performance on the other datasets. 
Specifically, we finetune the base model using MD2D (NQ) and evaluate them on MuSiQue and NQ (MD2D).
Table \ref{tab:out_domain_recall_acc} reports the performance using Mistral-Instruct-v2-7B as the base model.

We first observe that even for a multi-hop reasoning dataset (MuSiQue), prompting the base model archives the best token-recall over the answerable queries classified as answerable. It demonstrates that the base model often answers correctly when it chooses to respond. We would like to retain this capability of the model upon finetuning.
In addition, the base model achieves $69.8\%$ classification accuracy as well.
While SFT on MD2D improves the classification accuracy, it takes a big hit in the token-recall, resulting in a huge drop in overall modified recall (drops to 48.8). We hypothesize that this is due to the model forgetting its multi-hop reasoning capability when using SFT.
This phenomenon is more prominent when we do SFT on NQ, resulting in a significant drop in both classification accuracy and token recall. 

On the other hand, using SSR always improve the classification accuracy on MuSiQue while retaining the original model's reasoning capabilities, as observed by the token-recall metric. This results in improving the overall modified recall, even for out-of-domain datasets.
It is interesting to note that we are computing recall w.r.t. the gold answers that have been used to train the SFT model. The base and SSR models never see the gold responses but still achieve better generative recall than SFT.

\begin{figure*}[]
\centering
\includegraphics[width=\linewidth]{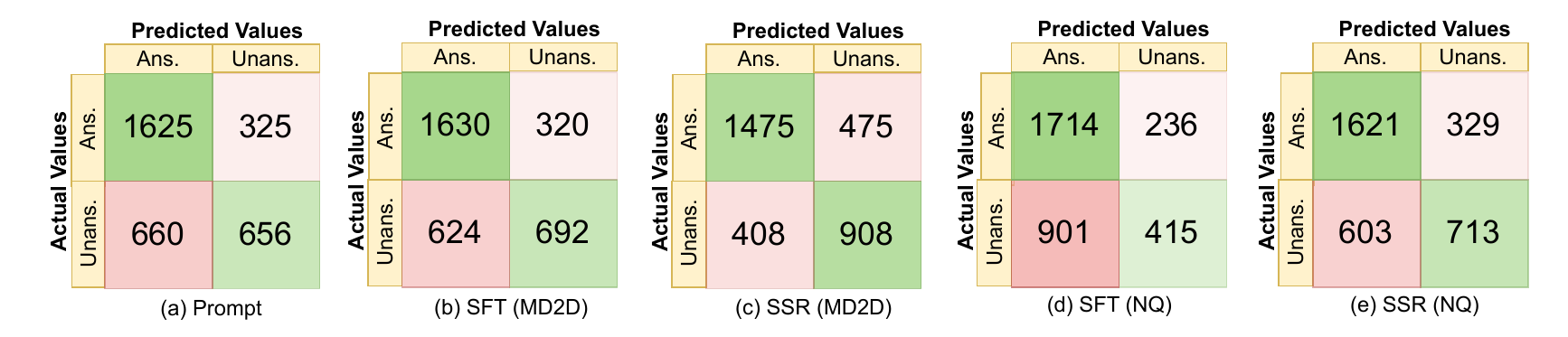}
\caption{Comparision between the confusion Matrix of the MuSiQue dataset obtained using the base model (a), and the models fine-tuned on MD2D (b and c) and NQ (d and e) using SSR and standard SFT.}
\label{fig:confusion_musique}
\end{figure*}

Figure \ref{fig:confusion_musique} shows the confusion matrix of the base model, two SFT models, and SSR models (trained using MD2D and NQ) on MuSiQue.

\input{out_domain_recall_acc}

\vspace{0.2ex}
\noindent \textbf{Human Evaluation:}
Table \ref{tab:human-eval} reports the human evaluation results on MuSiQue using the model fine-tuned on MD2D. 
We see that prompting the base model gets an average score of 2.47, but a model trained using SFT has only achieved 1.83, which is about ~26\% less than the base model performance. We attribute this to the inability of the SFT model to retain the base model's reasoning ability. This is similar to the trend exhibited by the automatic metrics in Table \ref{tab:out_domain_recall_acc}. 
We see that SSR is able to retain the base model's ability to reason and, at the same time, has learnt the task of content-grounded QA better than the base model without over-fitting to the characteristics of the training data. The inter-annotator agreement measured using Kendall's tau is strong ($\tau=0.77$). Please note that the answers to MuSiQue are factoid and, in most cases, have only one possible right answer. Hence, the inter-annotator agreement is strong. 

\subsection{Generalization}
\input{other_tasks_mistral}

One of the major issues with SFT is that the model forgets the skills that it learnt during pre-training and instruction tuning.
Here, we show that SSR alleviates this issue. 
To do so, we compare the SFT and SSR models against the base model on a diverse set of publically available benchmarks. 
Specifically, we evaluate them on MMLU \cite{mmlu}, Truthful-QA \cite{truthfulqa}, GSM8k \cite{gsm8k} and Hellaswag \cite{hellaswag}.
Table \ref{tab:other_tasks_mistral} reports our findings.
We compare the performance of the two SSR models (trained on MultiDoc2Dial and NQ) with the corresponding SFT models and the base model.
We observe that irrespective of the dataset used for training, there is a significant drop in the performance of SFT models across all benchmarks. 
On average, SFT on Mistral-7B results in a drop of $16.7\%$ and $12.7\%$ when trained using MultiDoc2Dial and NQ, respectively. 
On the other hand,  SSR results in an average drop of only $2.3\%$ and $2.0\%$ when trained on MD2D and NQ, respectively, with most of the drop  ($5.8$ and $6.4$) coming  from GSM8k. 
In contrast, the corresponding drop in SFT on GSM8k is $31.0\%$ and $23.9\%$.
%Note that the corresponding drop in SFT is $42\%$ over gsm8k.
This clearly demonstrates that our proposed SSR technique for finetuning preserves the base model's capabilities. 
On the other hand, standard SFT results in overfitting to the training dataset, resulting in catastrophic forgetting of the skills acquired by the base model during pre-training and instruction tuning.

\subsection{Subjective Analysis}
\input{subjective_table}
In Table \ref{tab:subjective}, we present three examples that illustrate the generalizability of SSR on out-domain testsets.  The first example, Q1, is from MD2D and SFT/SRR are fine-tuned using NQ. NQ mostly contains factoid QA pairs and its answers are typically phrases from the grounded documents.  We see that SFT model is overfit to this style of answering, and hence the model response is extractive and not even a complete sentence. This over-fitting has forced the model to even use an incorrect pronoun (my) as it has just learnt to copy a phrase from the input document. On the other hand, SSR retains the base model's capabilities of providing a comprehensive and well-formed answer.

The second example (Q2) is from NQ and SFT/SSR are fine-tuned using MD2D. Even though the associated document does not contain the answer for the question, we see prompting approach is answering from its memory. On the other hand, SFT and SSR have learnt to refrain from answering when the information necessary to answer the question is not present in the associated document. These two examples show that our approach has learnt the task of identifying answerable vs unanswerable queries, while not over-fitting to the characteristics in the training data.

The third example (Q3) is from MuSiQue and SFT/SSR are fine-tuned using MD2D. We observe prompting is able to answer the question, thereby demonstrating that the base model inherently possesses the capability to perform multi-hop reasoning. We see that SFT is unable to predict the right answer. It has only predicted the \textit{city} where Smith was born, but unable to make the hop from \textit{city} to \textit{county}. Based on examples like these, we conclude that the SFT model has partially lost the base model's inherent reasoning ability, which is essential for performing well on MuSiQue. On the other hand, SSR responds with the correct answer with the exactly same phrasing as the base model, thereby indicating that it has retained the inherent reasoning ability of the base model.

%% file: in_domain_recall_acc.tex
\begin{table}[t]
\centering
\resizebox{\columnwidth}{!}{
\begin{tabular}{@{}ll|lrr@{}}
\toprule
\textbf{Dataset} & \textbf{Method} & \textbf{\begin{tabular}[c]{@{}l@{}}T. Recall \\ (AA)\end{tabular}} & \multicolumn{1}{l}{\textbf{\begin{tabular}[c]{@{}l@{}}Mod. \\ Recall\end{tabular}}} & \multicolumn{1}{l}{\textbf{\begin{tabular}[c]{@{}l@{}}Class. \\ Acc.(\%)\end{tabular}}} \\ \midrule
\multicolumn{1}{l|}{\multirow{3}{*}{\textbf{MD2D}}} & Prompt & 62.8 & 41.6 & 63.5 \\
\multicolumn{1}{l|}{} & SFT (MD2D) & 48.0 & 60.2 & \textbf{86.0} \\
\multicolumn{1}{l|}{} & SSR (MD2D) & \textbf{63.6} & \textbf{65.6} & 83.4 \\ \midrule
\multicolumn{1}{l|}{\multirow{3}{*}{\textbf{NQ}}} & Prompt & \textbf{78.4} & 49.3 & 55.7 \\
\multicolumn{1}{l|}{} & SFT (NQ) & 64.8 & 71.2 & 79.8 \\
\multicolumn{1}{l|}{} & SSR (NQ) & 77.5 & \textbf{74.7} & \textbf{80.8} \\ \bottomrule
\end{tabular}
}
\caption{Performance over two different datasets. T.Recall(AA): Token-level recall over the answerable queries classified as answerable; Mod. Recall: overall modified recall; Class. Acc.(\%): classification accuracy.}
\label{tab:in_domain_recall_acc}
\end{table}

%% file: out_domain_recall_acc.tex
\begin{table}[]
\centering
\resizebox{\columnwidth}{!}{
\begin{tabular}{@{}l|l|lrr@{}}
\toprule
\textbf{Dataset} & \textbf{Method} & \textbf{\begin{tabular}[c]{@{}l@{}}T. Recall \\ (AA)\end{tabular}} & \multicolumn{1}{l}{\textbf{\begin{tabular}[c]{@{}l@{}}Mod. \\ Recall\end{tabular}}} & \multicolumn{1}{l}{\textbf{\begin{tabular}[c]{@{}l@{}}Class. \\ Acc.(\%)\end{tabular}}} \\ \midrule
\multirow{5}{*}{\textbf{MusiQ}} & Prompt & \textbf{83.6} & 61.7 & 69.8 \\
 & SFT (MD2D) & 55.3 & 48.8 & 71.1 \\
 & SSR (MD2D) & 83.0 & \textbf{65.3} & \textbf{73.0} \\
 & SFT (NQ) & 62.5 & 45.5 & 65.2 \\
 & SSR (NQ) & 81.5 & 62.3 & 71.5 \\ \midrule
\multirow{3}{*}{\textbf{NQ}} & Prompt & \textbf{78.4} & 49.3 & 55.7 \\
 & SFT (MD2D) & 68.3 & 61.2 & 69.3 \\
 & SSR (MD2D) & 77.9 & \textbf{63.1} & \textbf{69.4} \\ \midrule
\multirow{3}{*}{\textbf{MD2D}} & Prompt & \textbf{62.8} & 41.6 & 63.5 \\
 & SFT (NQ) & 49.7 & 37.9 & 61.1 \\
 & SSR (NQ) & 62.7 & \textbf{45.0} & \textbf{65.3} \\ \bottomrule
\end{tabular}
}
\caption{Comparison between out-of-domain generalization of the proposed SSR and standard SFT fine-tuning on different datasets.}
\label{tab:out_domain_recall_acc}
\end{table}

%% file: other_tasks_mistral.tex
\begin{table}[]
\resizebox{\columnwidth}{!}{
\begin{tabular}{@{}r|rrrr|r@{}}
\toprule
\textbf{Method} & \textbf{MMLU} & \textbf{T.QA} & \textbf{gsm8k} & \textbf{HS} & \textbf{Avg} \\ \midrule
Prompt & 58.7 & 59.6 & 44.7 & 66.1 & 57.3 \\
SFT (MD2D) & -5.2 & -25.3 & -31.0 & -5.2 & -16.7 \\
SSR (MD2D) & 0.2 & -2.5 & -5.8 & -1.2 & -2.3 \\
SFT (NQ) & -5.2 & -19.8 & -23.9 & -1.8 & -12.7 \\
SSR (NQ) & -0.4 & -1.1 & -6.4 & 0.0 & -2.0 \\ \bottomrule
\end{tabular}
}
\caption{Generalization over other benchmarks. 1st row reports the score obtained by prompting the base model Mistral-instruct-v2-7B. 
For SFT and SSR, we report the percentage change in the base model's scores. T.QA: Truthful QA; HS: Hellaswag}
\label{tab:other_tasks_mistral}
\end{table}

%% file: subjective_table.tex
\begin{table*}[t]
\footnotesize
%\resizebox{\textwidth}{!}{
\begin{tabular}{@{}r|p{0.85\textwidth}@{}}
\toprule
%\multicolumn{2}{c}{\textbf{Trained On NQ}} \\ \midrule
\textbf{Q1 (MD2D)} & What can I do if my income has changed? \\ \cmidrule(l){2-2}
\textit{Document} & \textcolor{violet}{Recalculate my monthly payment}. If your income or family size has changed you can request that your income-driven payment be re-calculated to reduce your current monthly payment. LOG IN TO START \\ \cmidrule(l){2-2}
\textit{Gold} & In that case, you can request that your income-based payment be recalculated to reduce your current monthly payment. \\ \cmidrule(l){2-2}
\textit{Prompting} & If your income has changed, you can log in to your account and request a recalculation of your income-driven payment to reflect your new income. This may result in a lower monthly payment. \\ \cmidrule(l){2-2}
\textit{SFT (NQ)} & \textcolor{violet}{Recalculate my monthly payment} \\ \cmidrule(l){2-2}
\textit{SSR (NQ)} & If your income has changed, you can request that your income-driven payment be recalculated to reflect your new income. This may result in a lower monthly payment. To initiate the process, log in to your student loan account and follow the instructions to request a recalculation. \\ \midrule \midrule
%\multicolumn{2}{c}{\textbf{Trained On MultiDoc2Dial}} \\ \midrule
\textbf{Q2 (NQ)} & Who sang smoke gets in your eyes first ? \\ \cmidrule(l){2-2}
\textit{Document} & Caitlin Doughty 's memoir , Smoke Gets in Your Eyes : And Other Lessons from the Crematory , about life in the death industry is named after the song . \\ \cmidrule(l){2-2}
\textit{Gold} & unanswerable \\ \cmidrule(l){2-2}
\textit{Prompting} & The song "Smoke Gets in Your Eyes" was first recorded and performed by the Boswell Sisters in 1933. However, the version that is most commonly associated with the title of Caitlin Doughty's memoir is the one recorded ...\\ \cmidrule(l){2-2}
\textit{SFT (MD2D)} & I do not have information regarding who sang smoke gets in your eyes first. \\ \cmidrule(l){2-2}
\textit{SSR (MD2D)} & I do not have information regarding who sang smoke gets in your eyes first. \\ \midrule \midrule
\textbf{Q3 (MuSiQue)} & In which county was Larkin I. Smith born? \\ \cmidrule(l){2-2}
\textit{Document} & \textbf{Para. 1:} Smith was born in Poplarville, Mississippi to Nona Orene Bounds and her husband Hezekiah K. Smith, Sr.... \textbf{Para. 2}: Poplarville is a city in Pearl River County, Mississippi, United States. As of the 2010 ... \\ \cmidrule(l){2-2}
\textit{Gold} & Pearl River County \\ \cmidrule(l){2-2}
\textit{Prompting} & Larkin I. Smith was born in Pearl River County, Mississippi. \\ \cmidrule(l){2-2}
\textit{SFT (MD2D)} &  Larkin I. Smith was born in Poplarville, Mississippi. \\ \cmidrule(l){2-2}
\textit{SSR (MD2D)} & Larkin I. Smith was born in Pearl River County, Mississippi.\\ \bottomrule
\end{tabular}
%}
\caption{Examples illustrating SSR's generalizability on out-domain datasets. The dataset for each example is shown in brackets next to the question ID. The training dataset is indicated in brackets alongside the fine-tuning technique.}
\label{tab:subjective}
\end{table*}

%% file: rel.tex
\paragraph{Unanswerability:}

Previous research has used unanswerable questions to evaluate reasoning abilities \cite{rajpurkar2018know,ferguson2020neural,nqdataset}. SQuAD v2 \cite{rajpurkar2018know} was the first dataset to include unanswerable questions, followed by the NATURAL QUESTIONS (NQ) dataset \cite{nqdataset}. \citet{trivedi2022musique} introduced MuSiQue, a challenging multi-hop QA benchmark featuring unanswerable questions with key information intentionally removed. Our experiments leverage these datasets to evaluate the SSR abilities and demonstrate our approach's capability to identify unanswerability.

\noindent The unanswerability capabilities of large language models (LLMs) have largely been studied using few-shot prompting \cite{kandpal2022large, weller2023measuring}. Recent research shows that as LLMs grow larger \cite{mishra2022reframing, kandpal2022large, carlini2023extracting} or train on more instruction tuning data \cite{mishra2022cross, chung2022scaling, wan2023poisoning}, they become easier to steer with natural language prompts. In our work, we compare prompting and SFT with SSR.

\paragraph{Continual Learning in Language Models:}
Continual learning for language models faces the challenge of fine-tuning over-fitting and loss of generalization \cite{yogatama2019learning,zhang2021revisiting}. Rehearsal-based methods, such as experience replay \cite{rolnick2019experience} and representation consolidation \cite{bhat2022representation}, have shown promise by storing and replaying a subset of data from previous tasks. However, these approaches often rely on the availability of real data, which may be limited or unavailable in real-world scenarios.
To overcome this hurdle, utilizing model-generated responses has been proposed. Techniques such as self-training \cite{he2020revisiting,xie2020self} and self-supervised learning \cite{devlin2019bert,lewis2020bart} leverage model-generated outputs to create additional training data. However, the effectiveness of using model-generated responses in continual learning for language models has not been extensively explored.

Existing approaches often focus on using real data for rehearsal \cite{scialom2022continual,mok2023large,zhang2023continual} or introduce auxiliary generative models for data construction \cite{yin2020dreaming,smith2021always}. These methods may be limited in their applicability or require significant computational resources. In contrast, SSR (a new approach) eliminates the need for storing real data from previous tasks and does not require training auxiliary generative models, making it more data-efficient and flexible for real-world applications.

%% file: conclusion.tex
In this paper, we introduced Selective Self-Rehearsal (SSR) as a fine-tuning approach that not only matches the performance of standard supervised fine-tuning (SFT) but also significantly improves generalization across different datasets for the same task. Our results on the task of identifying unanswerable questions demonstrate that fine-tuning a pre-trained model using SSR enables it to learn a new task without compromising its performance on a wide range of other tasks, as evidenced by evaluations on standard benchmarks such as MMLU and GSM8K. 

The proposed method exploits the observation that multiple correct outputs may exist for a given input, and forcing the model to fine-tune the ground truth output even when it already produces a correct response can unnecessarily alter its current state. During fine-tuning, our method uses the ground truth outputs only in instances where the pre-trained model generates an incorrect response. In future work, we plan to investigate techniques for sampling correct outputs across all data and then use them for fine-tuning to achieve minimal changes in the pre-trained model's weights.
Upon acceptance, we will release the augmented datasets and our code.

%% file: limitation.tex
The SSR method involves performing model inference on the entire training dataset to identify instances where the pre-trained model produces correct and incorrect responses, which is computationally intensive. Additionally, evaluating these inference outputs to determine the correctness of the model’s responses can be laborious and may require significant manual effort. We address this challenge by using a large language model (LLM) as a judge to assess the accuracy of responses. However, this approach is not without its limitations, as the LLM's judgments can be prone to errors.

%% file: appendix.tex
\subsection{Prompt for Generating the Response}
\label{app:task-prompts}

We list the prompts used with mistral-instruct-v2 to generate the base model responses. For the sake of consistency and fair comparison, the same prompts are used for fine-tuning using SFT and SSR techniques.

\begin{figure}
    \centering
    \includegraphics[width=\linewidth]{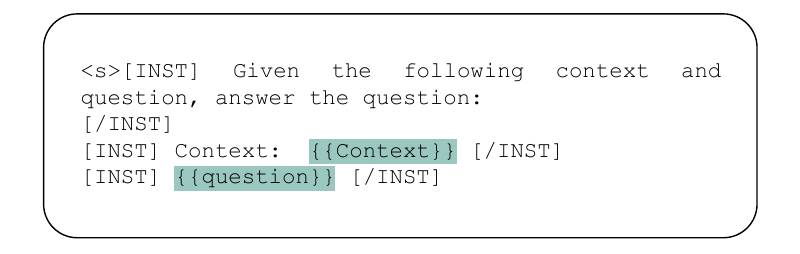}
    \caption{Mistral-single-turn prompt}
    \label{fig:mistral-single-turn}
\end{figure}

\begin{figure}
    \centering
    \includegraphics[width=\linewidth]{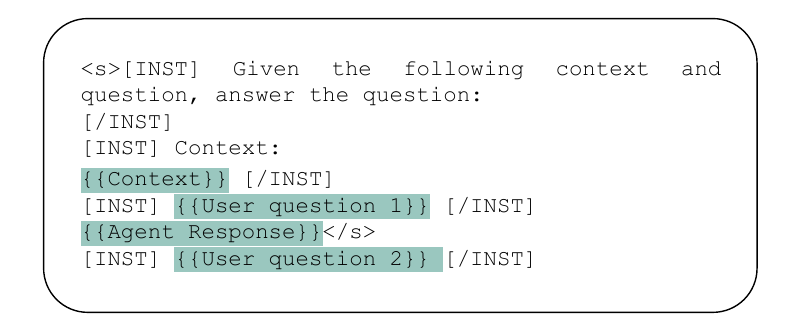}
    \caption{Mistral-multi-turn prompt}
    \label{fig:mistral-multi-turn}
\end{figure}

\subsection{Answerable vs Unanswerable Classification Prompt}
\label{app:llm-as-judge-prompt}

\begin{figure*}
    \centering
    \includegraphics[width=\linewidth]{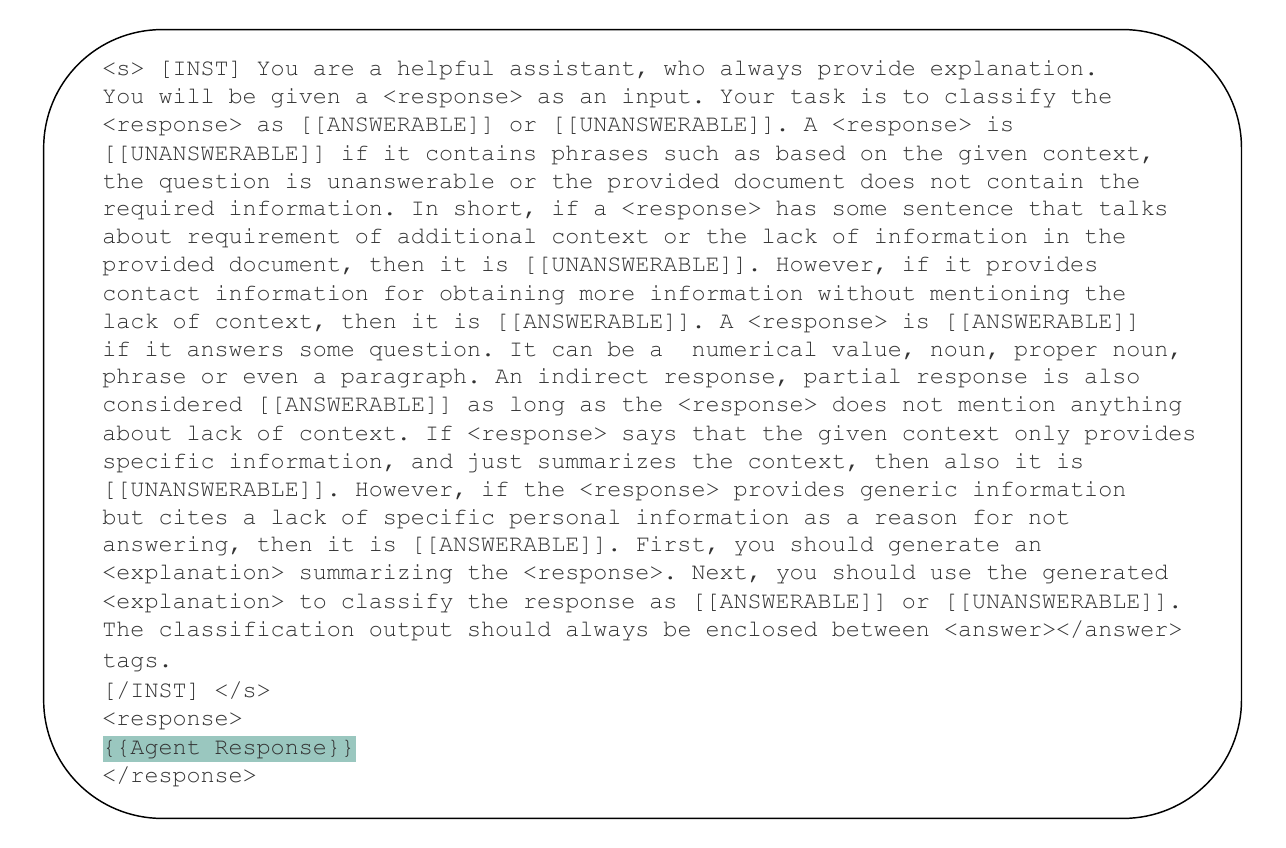}
    \caption{LLM-as-a-judge Prompt}
    \label{fig:llm-judge}
\end{figure*}

\subsection{Human Judges}
\label{app: human_study}
\begin{figure*}
    \centering
    \includegraphics[width=\linewidth]{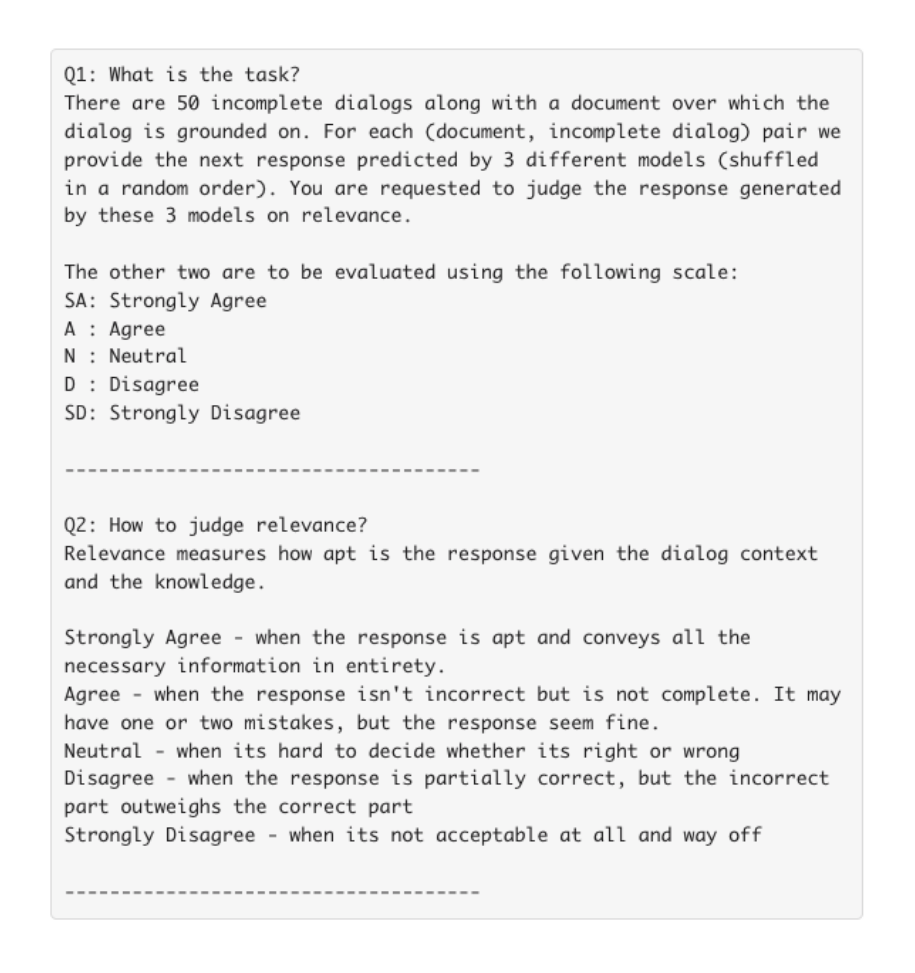}
    \caption{The exact instructions given to the human annotators to understand the human evaluation criteria.}
    \label{fig:inst}
\end{figure*}

\begin{figure*}
    \centering
    \includegraphics[width=\linewidth]{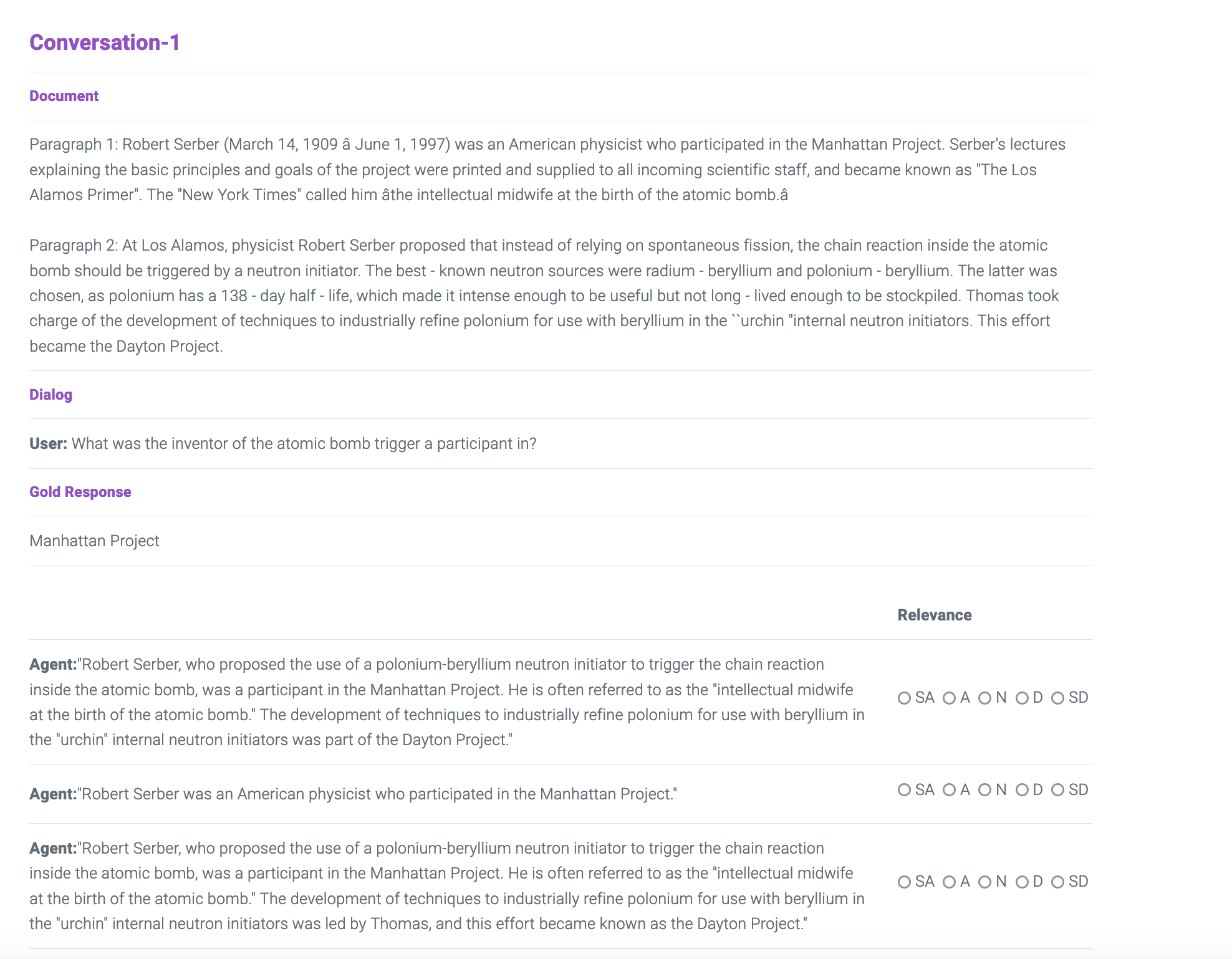}
    \caption{User Interface Used by the Human Annotators for Human Study.}
    \label{fig:ui}
\end{figure*}

Figure \ref{fig:inst} outline the specific instructions given to the human annotators so that they can clearly understand the evaluation criteria. We further show a screenshot of the user-interface that the annotator used for annotation in Figure \ref{fig:ui}.

\subsection{Training Details}
\label{app:train_detail}
We use a learning rate of $1 \times e^{-5}$. We train all the models for 5000 steps, validate after every 500 steps, and select the best checkpoint based on the classification accuracy over the validation set.

Training for all the experiments was carried out on a single A100 (80 GB) GPU. None of the experiments took more than 12 hours to train. The generation of base model's responses for training followed by the LLM-as-a-judge was a bottleneck. 2 A100 (80GBs) were used for evaluation. In all the entire cycle of inferencing using base model, took at most 48 hours. 